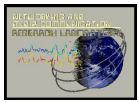

# A Study of Skews, Imbalances, and Pathological Conditions in LLM Inference Deployment on GPU Clusters detectable from DPU[1]

Javed I. Khan & Henry Uwabor Moye

August 2025
Media Communications and Networking Research Lab
Department of Computer Science
Kent State University, Kent OH 44242

### Abstract

*Autoregressive inference in large transformer-based language models (LLMs) presents significant challenges for runtime efficiency, particularly during the decode phase where load imbalance across GPU shards can cause throughput degradation and latency spikes. A DPU-assisted framework leveraged by BlueField-3 Data Processing Units can enable real-time detection and mitigation of load imbalance in multi-node tensor-parallel inference. By offloading monitoring tasks to the DPU and analyzing GPU telemetry and inter-node communication patterns, the resulting system can provide actionable feedback to inference controllers and schedulers. The goal of this study is three-fold i) identify the reported skews/imbalances/pathological conditions that arise in muti-GPU execution of a) LLM tensor computing (both during training and inference), b) identify their impact on computational performance, and c) make a critical assessment if those can be tracked for potential mitigation from a DPU network.*

## 1. Introduction

Transformer-based LLMs are widely deployed for natural language understanding and generation tasks [1]. However, their inference-time performance is increasingly constrained by compute and communication inefficiencies, particularly under tensor parallelism (TP) and pipeline parallelism (PP) [2]. While token batching improves average throughput, the decode phase often suffers from irregularities in token computation cost, leading to skew across parallel GPU workers. Existing software profilers (e.g., NVML, PyTorch hooks) add latency and lack token-level granularity. This paper explores the potential role of Data Processing Units (DPUs), specifically NVIDIA BlueField-3, in enabling lightweight, real-time observability and feedback for load imbalance detection [6].

## 2. Open Weight pretrained Models

Open-weight pre-trained models are large machine learning models—such as language models, vision transformers, or multimodal architectures—whose model weights are publicly released by the creators. Unlike closed models (e.g., OpenAI GPT-4, Anthropic Claude, Google Gemini), open-weight models allow developers to download the trained parameters, inspect them, fine-tune them on custom data, and redeploy them on their own hardware or in chosen inference environments.

The advantages of open-weight pre-trained models are substantial. First, they provide transparency and auditability: researchers can examine training data sources, architectures, and biases. Second, they enable customization and fine-tuning for domain-specific tasks—healthcare, legal text, multilingual services—without retraining from scratch. Third, open weights promote innovation and reproducibility: academic groups and startups can benchmark fairly and build upon shared baselines. Fourth, they give organizations deployment flexibility: models can run on-premises, in the cloud, or across specialized inference engines such as vLLM, Hugging Face TGI, or NVIDIA Triton, ensuring control over cost and performance. Finally, open weights lower barriers for global access, enabling small labs and emerging regions to participate in frontier AI development. In summary, open-weight pre-trained models democratize access to advanced AI while supporting efficiency, customization, and transparency in real-world deployments. Table-1 provides a sample list of such models. It's a growing list.

| Table- 1 Sample List of Open-Weight Pre-Trained Models for Redeployment | | | | |
|---|---|---|---|---|
| **Family** | **Sizes** | **Origin** | **Inference Engines** | **Application Domains** |
| LLaMA-2 / LLaMA-3 | 7B, 13B, 70B | Meta AI | vLLM, TGI, DeepSpeed, TensorRT, Triton, ORT | General-purpose LLMs; widely used for chat, research, fine-tuning, enterprise assistants |
| Mistral / Mixtral (MoE) | 7B (dense), 8×7B (MoE) | Mistral AI | vLLM, TGI, DeepSpeed, TensorRT, Triton | Efficient, strong reasoning; Mixtral MoE excels at scaling to large deployments |
| Falcon | 7B, 40B, 180B | TII (UAE) | vLLM, TGI, DeepSpeed, Triton, ORT | Optimized for efficiency & throughput; popular in enterprise and cloud serving |
| GPT-NeoX / GPT-J | 6B, 20B | EleutherAI | vLLM, TGI, DeepSpeed, Triton | Early open GPT-style models; used in research, prototyping, and academic projects |
| Pythia | 70M → 12B (multiple checkpoints) | EleutherAI | vLLM, TGI, DeepSpeed, Triton | Transparent scaling experiments; strong for benchmarks, interpretability, scaling law studies |
| OPT | 125M → 66B | Meta AI | vLLM, TGI, DeepSpeed, Triton | General-purpose baseline; often used for evaluation, benchmarking, and lightweight deployment |
| BLOOM / BLOOMZ | 560M → 176B | BigScience | vLLM, TGI, DeepSpeed, Triton, ORT | Multilingual LLMs; BLOOMZ tuned for cross-lingual chat, translation, global applications |
| Phi-2 / Phi-3 | 1.3B, 2.7B, 7B | Microsoft | vLLM, TGI, ORT | Compact and efficient; optimized for reasoning, code assistance, education tools |
| Gemma | 2B, 7B | Google DeepMind | vLLM, TGI, Triton | Small-scale but high-quality; targeted for safe deployment, consumer apps, teaching |
| Qwen / Qwen-VL | 1.8B → 72B | Alibaba Cloud | vLLM, TGI, Triton | Text + vision variants; good for multimodal tasks, bilingual apps, and chatbots |
| Yi | 6B, 34B | 01.AI | vLLM, TGI, Triton | High-quality bilingual training; strong at multilingual chat, reasoning, and coding |

## 3. Inference Engines

An **inference engine** is the runtime system that exIEecutes a trained model (like a large language model) to produce predictions or outputs. It is a software that i) loads a trained model (weights, architecture, ii) accepts inputs (e.g., text prompts, images, embeddings), iii) executes the model efficiently on available hardware (CPU, GPU, TPU, or DPU), and then iv) returns the outputs (e.g., generated tokens, classifications, embeddings).

Its core functions include model execution and running the forward pass of the neural network. In the background it optimizes to minimize latency and maximize throughput, performs memory and resource management, manages placement of model weights, activations, and KV-cache across CPU/GPU memory, handles batching, caching, and paging for efficiency. It also can be thought of as a hardware abstraction that provides a unified API for running on GPUs, CPUs, TPUs, or custom accelerators. It also uses libraries like CUDA, cuDNN, NCCL, ROCm, or oneDNN under the hood. Table-2 provides a survey of sample IEs.

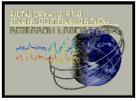

| Table-2 (a) Sample of Major Inference Engines | | | | | |
|---|---|---|---|---|---|
| **Engine** | **Key Features** | **GPU Scaling** | **Readiness** | **Pros** | **Cons** |
| **vLLM** | PagedAttention (KV-cache paging), continuous/dynamic batching, Hugging Face & OpenAI API compatibility | Multi-GPU supported (DP/TP), efficient memory reuse | Actively maintained, production-ready (cloud & on-prem) | High throughput, long-context support, efficient memory | Limited support for highly customized ops; younger ecosystem than Triton |
| **TGI (Text Generation Inference, Hugging Face)** | Optimized serving for Transformers, tensor/sequence parallelism, token streaming | Multi-GPU with DeepSpeed & Megatron integration | Production-grade, widely used in industry | Stable, easy deployment with Hugging Face hub, API ready | Less aggressive memory optimization vs. vLLM |
| **DeepSpeed-Inference (Microsoft)** | Kernel fusion, quantization (INT8, FP16, BF16), tensor parallelism, ZeRO inference | Scales across many GPUs with PP + TP | Production-ready, especially for MS ecosystem | Very efficient kernel optimizations, low-latency serving | Setup complexity, tied closely to PyTorch |
| **NVIDIA TensorRT / TensorRT-LLM** | Graph optimization, mixed precision kernels, CUDA Graphs, TensorRT runtime | Strong multi-GPU scaling (via NCCL, TP/PP support) | Highly production-ready, supported by NVIDIA ecosystem | Extremely optimized for NVIDIA GPUs, low latency | Vendor lock-in, limited portability to non-NVIDIA hardware |
| **ONNX Runtime (ORT)** | Supports many frameworks (PyTorch, TensorFlow, etc.), graph optimizations, quantization | Multi-GPU support improving, but less mature for LLMs | Production-ready, strong MS Azure integration | Broad framework support, portable across hardware | Slower for very large models vs. vLLM/TensorRT |
| **Ray Serve** | Scalable distributed serving, integrates with vLLM, TGI, and custom backends | Horizontal scaling across clusters | Production-ready for cloud-native deployment | Flexible, integrates into orchestration (Ray, K8s) | Overhead higher than engine-native serving |
| **Triton Inference Server (NVIDIA)** | Multi-framework support (PyTorch, TensorFlow, ONNX, vLLM backend), dynamic batching, monitoring | Multi-GPU and multi-node scaling | Enterprise-grade, widely used in HPC/AI serving | Unified deployment, strong observability, DPU integration | Configuration complexity, NVIDIA-focused |

## 3.1. Optimizations Services Offered by Inference Engines

Inference engines apply a wide range of software and hardware optimizations [3,4] to make the execution of trained models as fast and efficient as possible. Since inference often operates under strict latency and throughput requirements—serving real-time requests from thousands of users—the goal is to maximize performance while minimizing resource usage.

A first class of optimizations involves precision management. Instead of running models in full 32-bit floating point, inference engines exploit mixed-precision arithmetic (FP16, BF16, or even INT8). Quantization reduces memory footprint and improves compute throughput, while kernel fusion ensures numerical stability by merging multiple small operations into a single GPU kernel launch.

Another major optimization is efficient batching and scheduling. Engines like vLLM or TGI use dynamic or continuous batching to merge requests arriving at different times into shared GPU execution blocks, improving utilization. Some engines further apply length bucketing, grouping sequences of similar size to avoid stragglers. Together, these techniques reduce GPU idle time and improve fairness across user requests.

Memory management is equally important. Inference engines carefully manage KV-caches for transformer attention, embeddings, and activations. Techniques such as PagedAttention (in vLLM) [8] implement paging strategies similar to virtual memory systems, reusing or evicting cache blocks without wasting GPU memory. Memory preallocation and zero-copy transfers reduce overhead from repeated allocations.

On the hardware side, engines integrate with optimized backends like NVIDIA TensorRT, cuBLAS, cuDNN, or ROCm. These libraries provide highly tuned kernels for matrix multiplication, convolution, and attention primitives. Engines also exploit CUDA Graphs to amortize kernel launch overhead and use asynchronous streams to overlap compute with data transfers.

Finally, inference engines optimize the serving layer by enabling streaming outputs, RPC integration, and load-balancing across GPUs or nodes. This ensures not only fast token generation but also predictable latency distribution across users.

| Table -2 (b) Sample Real-Time Signals used Current Inference Engines | | | |
|---|---|---|---|
| **Signal** | **Origin (SW/HW)** | **Level** | **Use** |
| Request arrival time | SW (scheduler record keeping) | Application/server | Dynamic batching, admission control, |
| Sequence length | SW (tokenizer state) | Application/runtime | Length bucketing, batch formation, |
| Decode progress (# | SW (decoder state) | Application/runtime | Scheduling of decode iterations, |
| Queue depth / wait time | SW (inference engine queue) | Application/server | Admission control, tail-latency |
| KV-cache occupancy (pages in | SW (PagedAttention / cache tables) | Runtime/memory manager | Cache eviction, reuse, paging decisions |
| GPU utilization | HW counters via NVML/CUPTI | Device/GPU | Detect underutilization, |
| GPU memory | HW counters via CUDA/driver APIs | Device/memory | Prevent OOM, guide |
| PCIe / DMA throughput | HW counters via NVML/driver | System I/O | Detect host↔GPU congestion, |
| Kernel execution times | HW accessible (CUDA events / NVTX) | Device/runtime | Identify stragglers, phase profiling, CUDA Graph |
| Network queue depth / packet | HW (NIC/DPU telemetry) | Network stack | Detect jitter, microbursts, retransmits, egress |
| gRPC/QUIC request | SW (server transport layer) | Application/network | Admission control, backpressure |

### 3.2. How Engines Get the information?

Inference engines rely on a diverse set of real-time signals [5,9] to optimize performance under dynamic workloads [7]. At the request level, the system records arrival times to enable dynamic or continuous batching, while prompt lengths and tokenized sequence sizes guide length bucketing and predict memory requirements. Ongoing decode progress, measured in tokens already generated, informs when requests can be merged into batches without stalling others. On the GPU side, streaming multiprocessor (SM) utilization highlights whether devices are idle or overloaded, while kernel execution times obtained from CUDA events help detect stragglers and adapt scheduling. Monitoring memory bandwidth usage allows engines to decide when KV-cache paging or compression is needed. The memory state is tracked closely through VRAM usage metrics (via CUDA/NVML), KV-cache occupancy, and eviction or paging statistics such as those maintained by vLLM's PagedAttention, all of which are critical to avoid out-of-memory conditions during high concurrency. At the scheduling layer, the system maintains the number of pending requests per batch, queue depth, and wait times, using these to drive admission control and dynamically balance throughput against latency by adjusting batch sizes. Finally, at the serving layer, engines monitor gRPC/QUIC request latencies per client along with throughput and backpressure signals from NICs or DPUs; if hardware queues approach saturation, the inference server can throttle batching or slow acceptance of new requests to prevent overload. Together, these runtime signals form the feedback loop that enables inference engines to maximize GPU utilization while keeping user-facing latency under control. However, most of these real-time signals come from software-side record keeping and runtime instrumentation, not from "raw hardware snooping" (though some low-level counters do come from the GPU driver). Table-2(b) shows some sample signals, their origin, and their use.

## 4. Data Processing Units (DPU)

### 4.1. DPUs' Potential to Observe Network Pathologies

Because DPUs sit inline with the NIC, they process all ingress and egress traffic before it reaches the host CPU or GPU. This vantage point makes them uniquely positioned to observe network-level anomalies that impact distributed inference. DPUs can timestamp every packet with sub-microsecond accuracy, measure queue depth, burst rates, and flow irregularities, and export metrics without burdening the CPU. This enables early detection of microbursts—short traffic spikes that overflow switch buffers and introduce jitter—as well as persistent congestion that inflates token streaming latency.

DPUs also directly observe east–west traffic between GPUs across nodes, since all RDMA and NCCL collectives traverse NIC hardware. Even rare packet drops or transient congestion can create hidden retransmits and retries in QUIC/gRPC or RDMA transports, showing up only as unexplained latency spikes at the application layer. DPUs can detect these events in real time by tracking retransmission counters, duplicate ACKs, and queueing delays inside the NIC. Similarly, DPUs can expose egress serialization delays, where output tokens are generated by GPUs but stall in NIC queues waiting for line-rate transmission. This effect shifts the bottleneck from compute to network, inflating user-perceived latency even when GPUs are fully utilized.

By continuously monitoring these conditions, DPUs can provide fine-grained telemetry on packet loss, jitter, microbursts, and buffer congestion—issues that are otherwise invisible to CPUs or GPUs. In effect, DPUs transform from offload engines into network observability nodes, giving operators actionable insights to diagnose performance skew and stability problems in large-scale inference deployments.

## 4.2. DPUs' Potential to Observe Computational Skews

Beyond the network, DPUs can also interestingly expose imbalances inside the host node by monitoring communication between CPUs, GPUs, and memory over PCIe. Every host–device data transfer—embeddings, KV-cache writes/reads, or logits—travels as DMA transactions across the PCIe root complex. Because DPUs are PCIe peers, they can observe these transactions at high resolution. This allows DPUs to detect when PCIe saturation occurs, a common bottleneck in inference once embeddings or KV caches become large, and GPUs sit idle waiting for data.

DPUs can also watch for low-level GPU activity signals such as doorbell writes and CUDA stream identifier fetches. By correlating these with ingress request flows, a DPU can distinguish between prefill (large bursts of DMA writes) and decode (many small repetitive reads). This enables phase-level tracing of LLM inference workloads without modifying the application. When phase boundaries stretch abnormally—say, prolonged prefill DMA before compute begins—the DPU can flag potential CPU-side tokenization or batching bottlenecks.

In distributed setups, DPUs at each node can contribute to end-to-end skew detection. For example, if one GPU consistently exhibits delayed PCIe activity after ingress, the DPU can attribute the slowdown to local imbalance (e.g., CPU preprocessing lag, PCIe congestion) rather than network effects. Conversely, if PCIe patterns are healthy but responses stall at egress, the issue is likely network-side. This distributed view enables root-cause attribution: is skew introduced by host-to-GPU transfers, GPU scheduling, or external communication?

By combining fine-grained PCIe observability with correlation across multiple nodes, DPUs evolve into computational telemetry agents, capable of detecting skew at every step of the inference lifecycle and helping orchestrators dynamically rebalance workloads.

## 4.3. Computational Aspects DPUs Cannot See

Despite their unique vantage point on PCIe and network activity, DPUs cannot directly observe intra-GPU computation. Operations such as matrix multiplications, attention score calculation, or layer normalization happen entirely inside the GPU's execution pipelines and high-bandwidth memory (HBM/VRAM). These workloads never traverse PCIe or the NIC, so the DPU cannot monitor kernel-level utilization, arithmetic intensity, or memory-bank conflicts. Similarly, GPU-to-GPU collectives over NVLink/NVSwitch bypass the PCIe root complex, making them invisible to DPUs unless traffic spills over to PCIe or RDMA. The same limitation applies to CPU-only computations: preprocessing, tokenization, or intra-socket shared-memory exchanges are not observable unless they trigger PCIe or network transfers. In short, while DPUs excel at detecting communication bottlenecks—host↔GPU transfers, inter-node traffic, and NIC queueing—they cannot see into the black box of GPU kernels or CPU microarchitectural events. For complete observability, DPUs must therefore be paired with in-situ telemetry (e.g., GPU performance counters, CUDA profiling APIs, CPU PMCs) to bridge the gap between communication-level monitoring and actual compute execution.

## 4.4. Sample List of Skews/Imbalances/Pathological conditions in Tensor Computing

### Table-3 (a) North-South Runbook

| Skew /Imbalance | Signal (Red Flag) | Lifecycle Stages Affected | Effect on Node↔Node Traffic | Likely Root Cause | Mitigation Directives |
|---|---|---|---|---|---|

| **Burst admission backlog** | Sudden spikes of ingress requests ($T_0$ bursts) followed by queueing delay | **Ingress (prefill/start)** | Downstream GPU sees uneven load; internode bursts clump | Load spike from clients, front-end batching, NIC queue limits | Smooth input batching, rate-limit clients, increase NIC queue depth |
|---|---|---|---|---|---|
| **Ingress starvation / thin traffic** | Long gaps between ingress packets for some tokens | **Ingress → PCIe feed** | Token stalls; fewer collective ops downstream | Upstream service jitter, uneven client distribution | Balance load balancer hashing, check NIC RSS/flow steering |
| **Flow skew across sessions** | Some ingress flows high-volume, others sparse | **Ingress (per-request)** | Imbalanced TP/PP participation across tokens | Session affinity mismatch, QUIC stream imbalance | Verify flow hashing, rebalance RPC streams |
| **Ingress drop/ retransmit** | Missing or retransmitted initial packets (e.g., handshake retries) | **Ingress (request birth)** | Token ID not consistently assigned; lifecycle gaps | Congestion, MTU mismatch, link errors | Enable NIC offloads (TSO/GRO), verify MTU settings, check cabling |
| **Egress backlog / queueing** | Responses accumulate in NIC queues before send | **Egress (response flush)** | Downstream clients see latency spikes | CPU copy bottleneck, NIC buffer exhaustion | Offload checksums, use zero-copy send, increase NIC buffer size |
| **Egress jitter** | Outgoing packets for a token are spread unevenly over time | **Egress (decode outputs)** | Clients see irregular token cadence | Scheduler variance, CPU↔NIC contention | Isolate runtime threads, pin NIC IRQs, increase batching window |
| **Egress drop /retransmit** | Retransmissions or gaps in final response streams | **Egress** | Client-visible stalls; retries inflate latency | NIC offload misconfig, fabric congestion, buffer underrun | Check offload settings, enable congestion control (ECN/PFC) |
| **Early complet** | Some egress flows terminate far | **Egress (multi-** | Internode peers still busy; | Early-stop on short sequences; no | Enable inflight remapping / load stealing for decode |

| Skew/Imbalance | Signal (Red Flag) | Lifecycle Stages Affected | Effect on Node↔Node Traffic | Likely Root Cause | Mitigation Directives |
|---|---|---|---|---|---|
| **ion skew** | earlier than peers | **stream decode)** | imbalance in final stages | remap of freed resources | |
| **Ingress/ Egress bandwidth saturation** | NIC RX/TX at or near link capacity; queue buildup | **Ingress + Egress** | All internode phases elongated; cluster-level slowdown | Shared NIC with storage/other jobs; insufficient link | Upgrade NIC, QoS partitioning, stagger workloads |

## Table -3 (b) PCIe Observer Runbook

| Skew /Imbalance | Signal (Red Flag) | Lifecycle Stages Affected | Effect on Node↔Node Traffic | Likely Root Cause | Mitigation Directives |
|---|---|---|---|---|---|
| **H2D data starvation** | Large/clustered H2D DMAs followed by long gaps before doorbells/kernels | **Ingress→PCIe** (prefill & decode input feed) | Fewer/late internode bursts; downstream TP/PP idles | PCIe BW cap, NUMA miss, pageable (unpinned) host buffers | Pin memory, bind to correct NUMA socket, verify PCIe link width/speed |
| **D2H return-path bottleneck** | D2H DMAs linger / complete slowly; backlog after kernels | **Egress** (logits/tokens back to host) | Late responses; backpressure into next token step | PCIe saturation, IOMMU contention, CPU copy hotspots | Enable large pinned buffers, reduce copies, check IOMMU/ATS config |
| **Kernel launch/control latency** | Doorbells sporadic; long idle gaps between small H2D bursts and next launch | **Compute** (GPU underutilized across prefill/decode) | TP collectives delayed; PP handoffs drift | Runtime overhead, CPU scheduler delays, too many tiny kernels | Batch ops, fuse kernels, raise runtime launch queues, isolate CPU cores |
| **Intra-node GPU skew** | One GPU shows thin/irregular DMA; peers steady | **Compute** (per-layer) → propagates to **Internode** | TP collectives widen (straggler), PP stage misalignment | Uneven microbatching, memory pressure on a single GPU | Rebalance microbatches, unify stream priorities, check that GPU's memory and clocks |
| **PCIe link saturation** | Sustained near-peak PCIe throughput; | **Ingress→PCIe, Egress** | Burstiness in internode waves; | Oversubscribed PCIe switch / x8 link, competing | Verify x16 Gen/lanes, move devices off shared switch, stagger I/O |

| | compute stalls periodically | | elongates token step | DMAs (storage/NIC) | |
|---|---|---|---|---|---|
| **GPU P2P throttling (PCIe)** | P2P DMAs slow/variable; no NVLink path | **Compute** (intra-box TP/PP) | Internode timing jitter (collectives wait on slow intra-box move) | Shared uplink on PCIe switch; ACS/ATS settings | Prefer NVLink/NVSwitch; if PCIe, place GPUs under same switch, tune ACS/ATS |
| **Pinned-memory shortage / fragmentation** | Many small DMAs vs large coalesced; rising DMA count | **Ingress→PCIe** (feed) and **Egress** (returns) | Micro-jitter; uneven stage timing | Insufficient pinned pools; fallback to pageable | Pre-allocate larger pinned pools; coalesce transfers |
| **Host CPU bottleneck** | Low DMA rate despite available PCIe BW; delayed doorbells | **Compute orchestration** | Irregular TP cadence; PP bubbles | CPU contention, IRQ affinity, polling disabled | Isolate IRQs/threads, enable busy-poll where appropriate, pin runtime threads |
| **Memory registration churn** | Frequent map/unmap patterns around DMAs | **Ingress→PCIe** | Small timing gaps accumulating per token | Repeated registration due to short-lived buffers | Reuse registered buffers; RDMA/GPUDirect with persistent MR |
| **Decode early-stop skew** | D2H drops off early on some streams/GPUs | **Compute** (decode) → Egress | Some peers go silent; collectives wait for remaining peers | Sequence length variance; scheduler not rebalancing | Enable inflight request remapping/packing; speculative decode policies |

## Table- 3(c) East West Sensing RunBook

| Skew /Imbalance | Signal (Red Flag) | Lifecycle Stages Affected | Effect on Node↔Node Traffic | Likely Root Cause | Mitigation Directives |
|---|---|---|---|---|---|
| **TP straggler** | Wide arrival spread of collective bursts (max–min arrival gap ↑) | **Compute (Tensor Parallel collectives)** | Collective ops stall waiting for slowest peer | Skewed GPU load, PCIe starvation, memory imbalance on one node | Rebalance shards, check PCIe feeds per node, adjust affinity |
| **PP bubble / stage stall** | Large or growing gaps between stage handoff bursts | **Pipeline Parallel** | Downstream stage idles; upstream builds backlog | Load imbalance across pipeline stages, early | Adjust microbatch partitioning, reassign stages, speculative fill |

| | | | | token exit variance | |
|---|---|---|---|---|---|
| **Cross-node load skew** | Uneven traffic volume per node for same collective | **TP/PP compute → Internode** | Some nodes oversend/under send; throughput uneven | Shard imbalance, misaligned activation partitioning | Validate shard sizes, rebalance across nodes |
| **Network congestion / oversubscription** | Periodic spikes in latency + jitter across many links | **Internode transfers (collectives & stage handoff)** | Token step elongates cluster-wide | Fat-tree oversubscription, ToR link hot spot | Check fabric counters, enable adaptive routing, spread ranks |
| **Head-of-line blocking** | Some streams stall while others flow; out-of-order bursts | **Collective streams / P2P flows** | Latency-sensitive ops delayed | Shared queue depth exhaustion, RoCE/NIC queue imbalance | Increase NIC queue depth, enable QoS/ECN, verify fair sharing |
| **Retransmissions / packet loss** | Gaps + duplicate traffic or sudden retransmit storms | **All distributed phases** | Bursty latency; collectives jitter | Fabric errors, congestion collapse, misconfigured PFC | Verify lossless config, tune buffer thresholds, check optics/cabling |
| **Credit starvation (RDMA/ flow control)** | Long silence periods until remote credit update | **Internode (RDMA ops)** | Under-utilized links; token latency grows | Too-small RDMA window, NIC credit depletion | Increase QP window, tune flow control params |
| **KV-cache transfer bottleneck** | Repeated large bursts for some tokens, others silent | **Decode phase (PP handoff)** | Uneven memory pressure per stage; downstream skew | Sharded KV too large for link budget; non-uniform length | Compress KV, shard differently, apply caching policies |
| **Early-stop skew across nodes** | Some nodes stop sending mid-iteration while others continue | **Decode (multi-node)** | Collectives/pipeline hang waiting for peers | Sequence length divergence; scheduler not masking early exits | Enable dynamic remapping, mask early-stop ranks |

## 5. Conclusions

Mitigating computational and network skew in large-scale LLM inference requires fine-grained visibility into both application-level state and low-level system signals. Software-based sensing, embedded in inference engines such as vLLM, TGI, or DeepSpeed, provides rich metadata about requests, including arrival times, prompt lengths, decode progress, queue depth, and KV-cache occupancy. These records enable schedulers to perform optimizations such as dynamic batching, length bucketing, admission control, and cache eviction policies, ensuring that GPUs are fed efficiently while maintaining latency fairness across users. However, software-only sensing cannot see deeper system-level effects such as PCIe saturation, DMA burst contention, or NIC queueing, which often manifest as hidden bottlenecks. Here, DPU-based sensing complements the picture. Positioned inline with the NIC and as PCIe peers, DPUs can capture hardware timestamps, doorbell writes, DMA transaction sizes, jitter, microbursts, retransmits, and egress serialization delays. This allows DPUs to identify when GPUs are underutilized due to I/O congestion rather than poor batching, or when network anomalies inflate collective latency. By correlating software signals (e.g., request-level metadata, scheduling queues) with hardware signals (e.g., PCIe bursts, NIC buffer buildup), systems can achieve multi-layer observability. This hybrid view enables skew mitigation strategies such as rerouting requests away from congested nodes, dynamically resizing batches to fit available bandwidth, and triggering early KV-cache eviction when PCIe saturation is detected. The result is more precise root-cause attribution—distinguishing whether skew originates from CPU tokenization, GPU scheduling, PCIe bottlenecks, or network instability. Ultimately, combining software-based record keeping with DPU-based telemetry can create a much efficient closed feedback loop that would allow inference clusters to adaptively balance workloads, minimize idle bubbles, and deliver predictable low-latency performance at scale.

## 6. References

1. Dhakal, A., Kulkarni, S.G. & Ramakrishnan, K.K., 2023. *D-STACK: High throughput DNN inference by effective multiplexing and spatio-temporal scheduling of GPUs.* **arXiv preprint** arXiv:2304.13541. Available at: https://arxiv.org/abs/2304.13541 arXiv

2. Jain, P., Mo, X., Jain, A., Subbaraj, H., Durrani, R.S., Tumanov, A., Gonzalez, J. & Stoica, I., 2018. *Dynamic space-time scheduling for GPU inference.* **arXiv preprint** arXiv:1901.00041. Available at: https://arxiv.org/abs/1901.00041 arXiv

3. Razavi, K., Ghafouri, S., Mühlhäuser, M., Jamshidi, P. & Wang, L., 2024. *Sponge: Inference serving with dynamic SLOs using in-place vertical scaling.* **arXiv preprint** arXiv:2404.00704. Available at: https://arxiv.org/abs/2404.00704 arXiv

4.  Zhang, Z., Li, H., Zhao, Y., Lin, C. & Liu, J., 2023. *BCEdge: SLO-Aware DNN inference services with adaptive batching on edge platforms*. **arXiv preprint** arXiv:2305.01519. Available at: https://arxiv.org/abs/2305.01519 arXiv

5.  Langlet, J., Ben-Basat, R., Oliaro, G., Mitzenmacher, M., Yu, M. & Antichi, G., 2023. *Direct Telemetry Access*. In: *Proceedings of the ACM SIGCOMM 2023 Conference*, pp. 832–849. Available at: https://dl.acm.org/doi/10.1145/3603269.3604827 ACM Digital Libraryminlanyu.seas.harvard.edu

6.  Kfoury, E.F., Crichigno, J. & Bou-Harb, E., 2021. *An exhaustive survey on P4 programmable data plane switches: taxonomy, applications, challenges, and future trends*. *IEEE Access*, 9, pp. 87094–87155. Available at: https://doi.org/10.1109/access.2021.3086704 Scholar Commons

7.  Zhu, Y., Kang, N., Cao, J., Greenberg, A., Lu, G., Mahajan, R., Maltz, D., Yuan, L., Zhang, M., Zhao, B. & Zheng, H., 2015. *Packet-level telemetry in large datacenter networks*. In: *Proceedings of ACM SIGCOMM*. Available at: https://dblp.org/rec/conf/sigcomm/ZhuKCGLMMYZZZ15 dblp

8.  Arfeen, D., Zhang, Z., Fu, X., Ganger, G.R. & Wang, Y., 2024. *PipeFill: Using GPUs during pipeline-parallel LLM training*. **arXiv preprint** arXiv:2410.07192. Available at: https://arxiv.org/abs/2410.07192 arXiv.

9.  Javed I. Khan, Javed I., Thomas, P, Debabrata Das Robin, and Sharmila Rahman Prithula, *Real-Time Instrumentation and Telemetry Architecture for oVirt-Managed Virtual Machine Clusters for Campus Workload Characterization*, Technical Report 2025-07-01, Internetworking and Media Communications Research Laboratories, Department of Computer Science, Kent State University, available at: http://medianet.kent.edu/technicalreports.html